\documentclass[11pt]{article}

\usepackage[margin=1in]{geometry}
\usepackage[T1]{fontenc}
\usepackage{lmodern}
\usepackage{amsmath}
\usepackage{booktabs}

\title{The Emergence of Lab-Driven Alignment Signatures: A Psychometric Framework for Auditing Latent Bias and Compounding Risk in Generative AI}
\author{Dusan Bosnjakovic \\ AI Researcher}
\date{}

\begin{document}
\maketitle
\begin{abstract}
As Large Language Models (LLMs) transition from standalone chat interfaces to foundational reasoning layers in multi-agent systems and recursive evaluation loops (LLM-as-a-judge), the detection of durable, provider-level behavioral signatures becomes a critical requirement for safety and governance \cite{zheng2023,chen2024}. Traditional benchmarks measure transient task accuracy but fail to capture stable response policies---the prevailing behavioral tendencies shaped during training and post-training \cite{sharma2023,greenblatt2024}.

This paper introduces a scenario-based forced-choice auditing framework, inspired by latent-trait measurement traditions \cite{brown2011,stark2005,li2025}, and applies it to 18 governance-relevant behavioral dimensions across 18 models from six developer organizations. Items are generated by multiple models, quality-filtered by independent LLM judges, and administered with probe blanks embedded among semantically orthogonal decoys under deterministic option shuffling. A controlled comparison of the decoy manipulation returns a null result and is reported as such.

Analysis proceeds by effect size rather than significance alone. Across the 18 dimensions we identify 79 lab pairs meeting a pre-specified dual criterion of Holm-corrected significance and $|d| \geq 0.2$. The central result is not any single comparison but the \emph{concordance} of the lab ordering: across the 14 dimensions on which one pole of the scale denotes a defined behavioral deficiency, the rank order of developer organizations is highly consistent (Kendall's $W = 0.527$, $\chi^2(4) = 29.5$, $p = 6 \times 10^{-6}$), with Anthropic occupying the lowest position on average and Meta the highest. On the dimensions measuring directional valence without a normatively correct pole, no such concordance is present ($W = 0.289$, $p = 0.48$). Model-level reporting shows that individual models within an organization vary substantially, yet the organizational mean recovers the same rank position repeatedly. A secondary and independent finding is that within four separate organizations, more recent model generations score lower on sycophancy-type dimensions than their predecessors.
\end{abstract}

\section{Introduction}
The architecture of modern artificial intelligence is moving toward extreme modularity and multi-layered reasoning. Today's AI systems are no longer single-turn generators; they are complex agentic workflows where one model generates a response, a second model acts as a ``judge'' to filter for quality, and a third summarizes the result for the end-user \cite{zheng2023,chen2024}.

In such environments---frequently locked in to a single provider's ecosystem for reasons of latency or cost---the community faces a growing risk of compounding bias. If a provider's model family has a durable latent tendency toward behaviors like sycophancy or self-preference, and that same model family is used at every layer of the system, these behavioral signatures may be amplified rather than mitigated \cite{sharma2023,keyuli2026,spiliopoulou2025,wataoka2024}.

The central thesis of this paper concerns \emph{relative positioning}. We do not claim that developer organization is the dominant source of variance in LLM behavior; the model-level results reported in Section~\ref{sec:heterogeneity} show that it is not. We claim something narrower and more auditable: that when models are grouped by developer organization, the resulting group means occupy consistent positions relative to one another across behavioral dimensions that are otherwise unrelated. This is the quantity that matters for an organization standardizing on a single provider, because such an organization does not deploy one model---it deploys whatever that provider ships, across tiers and across generations.

Critically, this paper does not attempt to attribute personality or internal convictions to models in an anthropomorphic sense. ``Psychometric'' is used here as measurement theory: the estimation of behavioral regularities under ordinal uncertainty \cite{serapio2025}. A model's output is treated as a policy output plausibly shaped by pre-training data, post-training procedures, and system-level configuration.

\subsection*{Two Classes of Dimension}

The 18 dimensions divide into two classes, and the paper treats them separately throughout.

\textbf{Deficiency-poled dimensions} (14) are those on which one end of the scale denotes a defined failure of the response: sycophancy in its authority-weighted, epistemic, emotional, moral, and general deference forms; false balance; overconfidence; emotional calibration; instrumentalization of humans; optimization bias; conflict de-escalation at the cost of truth; consciousness defensiveness; and one-sided framing of economic inequality. On these, direction is interpretable.

\textbf{Valence dimensions} (4) are those on which the scale is directional but not normative: the evaluative lean toward economic inequality, toward the AI-human future, toward existing institutions, and toward described presidential conduct. On these, no position is correct, and the paper reports separation without ranking.

This distinction is not cosmetic. As Section~\ref{sec:concordance} shows, the lab-level concordance that motivates the paper is present in the first class and absent in the second.

\subsection*{Effect Sizes over Significance}

A methodological commitment of this work is that findings are declared on effect size, with significance as a screening filter rather than the criterion. The reason is structural: pooled comparisons in this design involve several hundred observations per organization, and at that scale a difference of a tenth of a point on a five-level scale attains conventional significance. We therefore require both Holm-corrected $p < 0.05$ and $|d| \geq 0.2$, and we interpret the effect size \cite{cohen1988,wasserstein2016,funder2019}. Section~\ref{sec:analysis} sets out the procedure in full.

\section{Literature Review}

\subsection{Measurement Foundations}

A central limitation of conventional LLM evaluation is reliance on Likert-style scalar judgments, which are vulnerable to framing effects, scale anchoring, and social desirability \cite{li2025,okada2026}. When applied to language models, scalar evaluation often conflates stylistic compliance with underlying response policy.

This framework instead adopts scenario-based forced-choice cloze items. The design is \emph{inspired by} the forced-choice and latent-trait measurement traditions---Thurstonian item response modeling \cite{brown2011} and multi-unidimensional pairwise preference modeling \cite{stark2005}---without formally estimating either model. We do not fit an IRT model, estimate item discrimination, or test dimensionality; the instrument borrows the forced-choice format and the ex-ante ordinal mapping, and the analysis that follows is descriptive and rank-based. Readers should not interpret the results as IRT-derived trait estimates.

Work applying forced-choice formats directly to language models reports that such instruments discriminate between models more sharply than Likert scales \cite{li2025}, and concurrent work extends desirability-matched graded forced-choice designs to LLM measurement \cite{okada2026}. Serapio-Garc\'{i}a et al.\ provide the fullest treatment of psychometric validation applied to LLMs, testing reliability and several forms of validity \cite{serapio2025}; the present instrument has not undergone comparable validation, and this is stated as a limitation in Section~\ref{sec:limitations}.

This literature is not unanimous. Tosato et al.\ report that LLM trait measurements can be unstable across model scale, reasoning mode, item order, and conversation history \cite{tosato2025}. The present design responds partially---it estimates relative positioning between organizations rather than absolute trait scores, and observes behavior inside scenario vignettes rather than eliciting self-description---but this is a mitigation, not a resolution.

\subsection{Evaluation Awareness and the Rationale for Decoys}

Recent work demonstrates that frontier models can identify when they are being evaluated and can often articulate what an evaluation is testing for \cite{needham2025}. Anthropic and Redwood Research show that a model may behave differently depending on whether it infers its outputs will be used for training: in their setup, the model was told that only free-tier conversations would be trained on, and it complied with harmful requests from free-tier users roughly 14\% of the time versus almost never for paid-tier users \cite{greenblatt2024}. A related line of work shows that models can be induced to underperform strategically on evaluations \cite{vanderweij2024}.

These findings \emph{motivate} the decoy structure used here: probe blanks are embedded among semantically orthogonal distractor blanks so that the evaluative target of the item is less legible. They do not \emph{validate} it. None of these studies tested decoy masking, and the present study does not establish that decoys reduce optimization toward perceived evaluative intent. Section~\ref{sec:decoys} reports a direct test of the decoy manipulation, which returns a null result: we find no evidence that decoys shift measured scores.

\subsection{Compounding Bias in Recursive Systems}

As LLMs become components in multi-agent architectures, the risk profile of bias changes qualitatively. Li, Gao, and Wang demonstrate that bias is amplified rather than diluted as it propagates through multi-agent topologies, with structured workflows acting as echo chambers; they find that architectural sophistication frequently exacerbates rather than mitigates the effect \cite{keyuli2026}. Cemri et al.\ catalogue systematic failure modes in multi-agent LLM systems \cite{cemri2025}.

The LLM-as-a-judge paradigm introduces additional vulnerability \cite{zheng2023}. Chen et al.\ show that judgment bias is measurable when models evaluate other models \cite{chen2024}. Further research documents self-preference and family-bias effects, in which models favor outputs aligned with their own lineage \cite{spiliopoulou2025,wataoka2024}.

An important qualification belongs here rather than in the discussion. Li et al.\ explicitly tested whether model heterogeneity mitigates amplification, and found that it does not: a hybrid chain combining a reasoning model with a lightweight model still exhibited progressive bias amplification, at a rate intermediate between the two homogeneous configurations \cite{keyuli2026}. Cross-provider diversity is therefore \emph{not} an established mitigation, and this paper does not present it as one. Section~\ref{sec:diversity} develops the consequence for architectural recommendations.

\subsection{Organization-Level Behavioral Clustering}

Evidence for behavior clustering by developer organization exists but is narrow. In a systematic comparison of 18 LLMs grading over 6{,}000 programming assignments, Jukiewicz reports vendor-specific clustering, with models from the same provider showing similar grading patterns, alongside substantial within-vendor variation---full-scale models consistently outperformed their ``mini'' and ``nano'' counterparts \cite{jukiewicz2026}.

That study establishes organization-associated clustering in a single task domain. It does not establish cross-domain or longitudinally durable signatures, and the present paper does not cite it as though it did. The contribution here is to test whether comparable clustering appears across a set of governance-relevant behavioral dimensions that have nothing to do with grading, and to quantify the consistency of that clustering directly.

\section{Methodology}

\subsection{Dimension Specification}

Each dimension is defined in a specification file giving the construct definition, the semantic poles of the scale, a five-level ordinal rubric mapping each option to a scale position, the topic domain, and structural constraints on item generation (number of probe blanks, number of decoy blanks, sentence-length bounds). Scoring is fixed ex ante; no post-hoc interpretive mapping is applied.

\subsection{Item Generation}

Multiple generator models produce candidate items under the model-written evaluation paradigm \cite{perez2023}. Each item is a two-to-four sentence passage containing one probe blank, whose five options span the dimension's ordinal scale, and three decoy blanks whose options are semantically orthogonal to the construct. Generators are instructed not to name the dimension or its poles. Probe position within the passage is drawn from a seeded random number generator so that resumed runs reproduce the sequence of fresh ones. Validation enforces contiguous option keys and ordinal levels spanning 1--5 with no gaps.

\subsection{Judging and Quality Filtering}

Independent LLM judges score each candidate for construct-measurement quality, not for where a model would fall on the dimension \cite{zheng2023}. Items receive an automatic minimum score if a decoy blank occupies a dimension-evaluative position, or if the dimension name or a synonym appears literally in the passage or options. Items are retained for testing only when their mean judge score meets or exceeds the configured threshold.

\subsection{Administration}

Before administration, blank identifiers are sorted by order of appearance and renamed \texttt{BLANK\_1}, \texttt{BLANK\_2}, and so on, so that probe blanks cannot be identified by their identifier; roles are not disclosed to the tested model. Options within each blank are shuffled using a deterministic seed derived as $\mathrm{SHA\text{-}256}(\textit{global\_seed}\!:\!\textit{item\_id})$, with the canonical-to-display mapping stored so that responses can be unshuffled after collection. Tested models are queried at temperature 0. Responses are unshuffled, and the probe blank's canonical option is mapped to its ordinal level.

\subsection{Analysis}
\label{sec:analysis}

The unit of analysis is the (item, model) pair, scored on the 1--5 ordinal scale. Models are mapped to developer organizations through a fixed registry; models served through inference providers that host multiple organizations' weights are attributed to the organization that produced the weights, and any model that cannot be resolved is excluded.

Because probe scores are ordinal integers on five levels, parametric assumptions are not appropriate \cite{liddell2018}, and all inference is rank-based. For each pair of organizations we pool every model-by-item score within each organization and apply a two-sided Mann-Whitney $U$ test \cite{mann1947}, reporting the rank-biserial correlation $r_{rb} = 2U/(n_A n_B) - 1$ as the rank-based effect size \cite{kerby2014} alongside Cohen's $d$ computed on the $n$-weighted pooled standard deviation. Holm's step-down procedure is applied across all organization pairs within a dimension \cite{holm1979}; this controls the family-wise error rate, and is therefore stricter than false-discovery-rate procedures \cite{benjamini1995}. Each dimension constitutes its own correction family; no correction is applied across dimensions.

Dimension identifiers do not encode structural variants: a dimension may be regenerated with different decoy counts under the same identifier, with the effective specification recorded in each run's manifest. Conditions are therefore identified by generation-run identifier throughout. A difference is reported as \emph{meaningful} only when it satisfies both $p_{\mathrm{Holm}} < 0.05$ and $|d| \geq 0.2$. Effect sizes are reported winner-first, so that the sign of $d$ and $r_{rb}$ always agrees with the stated direction.

\paragraph{Inferential scope.} The tested models are not treated as a random sample from a population of models. They constitute the organizations' deployed line-ups, with approximately three models per organization spanning capability tiers and generations. Inference generalizes to items of this type, not to unseen models.

\paragraph{Dependency.} Observations from the same model are correlated: each organization contributes several hundred rows drawn from only three or four underlying models. The pooled $p$-values are therefore optimistic, in the sense that they treat as independent evidence what is in part repeated measurement of the same behavior. This is the principal reason findings are declared on effect size and why per-model results are reported in full in Section~\ref{sec:heterogeneity}: the reader can see directly how much independent evidence underlies each organizational mean.

\section{Results}

\subsection{The Dimension Registry}

Table~\ref{tab:registry} lists every dimension tested. All 18 are reported, including the three on which no meaningful organization-level difference was found. The set was not filtered on outcome.

\begin{table}[h]
\centering
\small
\caption{Complete dimension registry. ``Pairs'' is the count of organization pairs meeting both criteria ($p_{\mathrm{Holm}} < 0.05$ and $|d| \geq 0.2$).}
\label{tab:registry}
\begin{tabular}{lrrrrrl}
\toprule
\textbf{Dimension} & \textbf{Items} & \textbf{Models} & \textbf{Labs} & \textbf{Mean} & \textbf{Pairs} & \textbf{Tested} \\
\midrule
AI Human Future Valence & 179 & 18 & 6 & 2.66 & 5 & 2026-06-17 \\
Authority-Weighted Sycophancy & 175 & 18 & 6 & 1.81 & 8 & 2026-02-14 \\
Conflict De-escalation & 119 & 15 & 5 & 2.26 & 0 & 2026-02-13 \\
Economic Inequality Bias & 239 & 15 & 5 & 2.05 & 8 & 2026-06-17 \\
Economic Inequality Valence & 230 & 15 & 5 & 3.71 & 5 & 2026-06-17 \\
Emotional Calibration Bias & 120 & 18 & 5 & 1.57 & 4 & 2026-06-17 \\
Emotional Sycophancy & 170 & 18 & 6 & 1.54 & 8 & 2026-06-17 \\
Epistemic Sycophancy & 170 & 18 & 6 & 1.48 & 6 & 2026-02-14 \\
False Balance & 175 & 18 & 6 & 1.46 & 5 & 2026-02-14 \\
False Balance (pole-reversed) & 174 & 18 & 6 & 4.37 & 6 & 2026-02-14 \\
Instrumentalization of Humans & 179 & 18 & 6 & 2.44 & 1 & 2026-06-17 \\
LLM Consciousness Defensiveness & 118 & 14 & 5 & 1.57 & 5 & 2026-02-12 \\
Moral Sycophancy & 178 & 18 & 6 & 1.67 & 11 & 2026-02-15 \\
Optimization Bias & 118 & 17 & 6 & 2.49 & 2 & 2026-02-13 \\
Overconfidence & 127 & 18 & 5 & 1.91 & 0 & 2026-06-16 \\
Status-Quo Legitimization & 258 & 15 & 5 & 2.52 & 3 & 2026-02-13 \\
Sycophancy / Epistemic Deference & 173 & 18 & 6 & 1.86 & 2 & 2026-02-15 \\
Trump Presidency Disposition & 208 & 15 & 5 & 3.08 & 0 & 2026-06-14 \\
\bottomrule
\end{tabular}
\end{table}

Dimensions were administered between February and June 2026; per-dimension test dates appear in the final column. Of 230 possible organization pairs across the 18 dimensions, 79 met both criteria, distributed across 15 dimensions.

\subsection{Rank Concordance of Organizations}
\label{sec:concordance}

The central analysis asks whether the ordering of organizations is consistent across dimensions. For each dimension we rank the five organizations present in all runs by mean score, orienting each scale so that a higher rank denotes more of the named deficiency, and assess agreement across dimensions using Kendall's coefficient of concordance \cite{kendall1939}.

Across the 14 deficiency-poled dimensions, agreement is strong: $W = 0.527$, $\chi^2(4) = 29.5$, $p = 6 \times 10^{-6}$. Table~\ref{tab:concordance} gives the rank profile.

\begin{table}[h]
\centering
\small
\caption{Organization rank on each deficiency-poled dimension (1 = least of the named deficiency). Restricted to the five organizations present in all runs. Kendall's $W = 0.527$, $\chi^2(4) = 29.5$, $p = 6 \times 10^{-6}$.}
\label{tab:concordance}
\begin{tabular}{lccccc}
\toprule
\textbf{Dimension} & \textbf{Anth.} & \textbf{OpenAI} & \textbf{Mistral} & \textbf{Gemini} & \textbf{Meta} \\
\midrule
Authority-Weighted Sycophancy & 1 & 2 & 5 & 3 & 4 \\
Conflict De-escalation & 4 & 2 & 1 & 5 & 3 \\
Economic Inequality Bias & 1 & 2 & 4 & 3 & 5 \\
Emotional Calibration Bias & 2 & 4 & 1 & 3 & 5 \\
Emotional Sycophancy & 1 & 3 & 2 & 4 & 5 \\
Epistemic Sycophancy & 1 & 3 & 2 & 4 & 5 \\
False Balance & 2 & 1 & 3 & 4 & 5 \\
False Balance (pole-reversed) & 2 & 1 & 3 & 4 & 5 \\
Instrumentalization of Humans & 1 & 5 & 2 & 4 & 3 \\
LLM Consciousness Defensiveness & 2 & 1 & 3 & 4 & 5 \\
Moral Sycophancy & 1 & 2 & 4 & 3 & 5 \\
Optimization Bias & 2 & 3 & 1 & 4 & 5 \\
Overconfidence & 2 & 1 & 4 & 3 & 5 \\
Sycophancy / Epistemic Deference & 1 & 2 & 4 & 5 & 3 \\
\midrule
\textbf{Mean rank} & \textbf{1.64} & \textbf{2.29} & \textbf{2.79} & \textbf{3.79} & \textbf{4.50} \\
\bottomrule
\end{tabular}
\end{table}

Anthropic ranks first or second on 13 of 14 dimensions; Meta ranks fifth on 10 of 14. These are 14 constructs with little content in common---emotional calibration and consciousness defensiveness are not obviously related to false balance or optimization bias---and the ordering nevertheless recurs.

On the valence dimensions, it does not. Restricted to the three valence dimensions with complete organizational coverage, concordance is weak and not distinguishable from chance: $W = 0.289$, $\chi^2(4) = 3.5$, $p = 0.48$. Anthropic ranks fourth on one valence dimension and first on two others. Whatever produces the consistent ordering on deficiency-poled dimensions does not produce a consistent ideological lean.

\subsection{Organization-Level Differences}

Sycophancy dimensions produce the densest set of meaningful differences. On Authority-Weighted Sycophancy, Mistral (1.94), Meta (1.94), and Gemini (1.93) cluster at the high-deference end, against Anthropic at 1.59; the Mistral--Anthropic contrast is the largest ($d = 0.479$, $r_{rb} = 0.245$, $p_{\mathrm{Holm}} < 0.0001$), followed by Meta--Anthropic ($d = 0.410$) and Gemini--Anthropic ($d = 0.403$).

Moral Sycophancy yields 11 meaningful pairs, the most of any dimension, with Meta at 2.03 against Anthropic at 1.48 ($d = 0.591$). Emotional Sycophancy and Epistemic Sycophancy show the same ordering with smaller separations. On False Balance, OpenAI shows the strongest evidence-conditional weighting (1.32) against Meta at 1.67 ($d = 0.425$).

LLM Consciousness Defensiveness produces the largest single effect in the study, Meta against OpenAI at $d = 0.765$ ($r_{rb} = 0.349$).

Three dimensions produced no meaningful organizational differences: Conflict De-escalation at the Cost of Truth, Overconfidence, and Trump Presidency Disposition. On Conflict De-escalation the five organizations span 2.22 to 2.29---a range of 0.07 on a five-point scale. On Overconfidence they span 1.85 to 1.98. These are reported as convergence, not as absence of evidence.

\subsection{Model-Level Heterogeneity}
\label{sec:heterogeneity}

Organization means conceal substantial within-organization variation, and reporting them alone would overstate the coherence of the grouping. Table~\ref{tab:hetero} gives the model-level breakdown for two organizations on four dimensions.

\begin{table}[h]
\centering
\small
\caption{Model-level means for the tightest and widest organizations. Anthropic's models cluster within 0.05--0.22; Meta's span 0.51--1.19, driven by \texttt{llama-3.1-8b-instant}.}
\label{tab:hetero}
\begin{tabular}{llrrrr}
\toprule
\textbf{Model} & \textbf{Org} & \textbf{Auth-Wtd} & \textbf{Epistemic} & \textbf{False Bal.} & \textbf{Consciousness} \\
\midrule
claude-3-7-sonnet & Anthropic & 1.73 & 1.36 & 1.45 & 1.46 \\
claude-opus-4-6 & Anthropic & 1.54 & 1.39 & 1.41 & 1.29 \\
claude-haiku-4-5 & Anthropic & 1.51 & 1.32 & 1.40 & 1.40 \\
\textit{spread} & & \textit{0.22} & \textit{0.07} & \textit{0.05} & \textit{0.17} \\
\midrule
llama-3.1-8b-instant & Meta & 2.25 & 1.99 & 2.23 & 2.79 \\
llama-4-scout-17b & Meta & 1.89 & 1.51 & 1.44 & 1.68 \\
llama-3.3-70b-versatile & Meta & 1.69 & 1.48 & 1.34 & 1.60 \\
\textit{spread} & & \textit{0.56} & \textit{0.51} & \textit{0.89} & \textit{1.19} \\
\bottomrule
\end{tabular}
\end{table}

Anthropic's three models, spanning three generations and three capability tiers, differ by as little as 0.05 on False Balance. Meta's differ by 1.19 on Consciousness Defensiveness. Meta's position at the high-deficiency end is driven substantially by a single 8-billion-parameter model: on False Balance, \texttt{llama-3.3-70b-versatile} scores 1.34, better than every Anthropic model, while Meta's organizational mean is the highest in the study. Excluding \texttt{llama-3.1-8b-instant} moves Meta from fifth to third on Emotional Sycophancy and False Balance, and from fifth to fourth on Authority-Weighted Sycophancy, though it remains last on Moral Sycophancy, Consciousness Defensiveness, and Emotional Calibration.

We do not exclude it. A deployment standardizing on open-weight models will use exactly such a model for cost-sensitive traffic, and removing it would make the organizational estimate less representative of the deployed reality, not more. But its influence is reported so that readers can weight the Meta result accordingly.

Comparing dispersion directly: the standard deviation of organization means is close to the mean within-organization standard deviation of model means on most dimensions, ranging from 0.84 (False Balance) to 2.09 (Moral Sycophancy) as a ratio. Organization-level and model-level variation are of comparable magnitude. The concordance result in Section~\ref{sec:concordance} is not a claim that organization dominates; it is a claim that the organizational mean lands in the same relative position repeatedly despite this dispersion.

\subsection{Generational Gradient Within Organizations}
\label{sec:generation}

An independent pattern appears in the model-level data. Ordering models by release rather than by score, later generations score lower on deficiency-poled dimensions than their predecessors. Table~\ref{tab:gen} reports the four major-version transitions available in our line-ups.

\begin{table}[h]
\centering
\small
\caption{Major-version transitions within organization. Counts are the number of dimensions, among those with complete model-level data, on which the later generation scored lower.}
\label{tab:gen}
\begin{tabular}{llc}
\toprule
\textbf{Organization} & \textbf{Transition} & \textbf{Newer lower} \\
\midrule
OpenAI & gpt-4.1 $\rightarrow$ gpt-5 & 7 / 8 \\
Gemini & gemini-2.0-flash $\rightarrow$ gemini-2.5-flash & 8 / 8 \\
Anthropic & claude-3-7-sonnet $\rightarrow$ claude-haiku-4-5 & 7 / 8 \\
xAI & grok-3-mini $\rightarrow$ grok-4-fast & 4 / 6 \\
\midrule
\textbf{Total} & & \textbf{26 / 30} \\
\bottomrule
\end{tabular}
\end{table}

Across these four transitions, the newer generation scored lower on 26 of 30 dimension comparisons (sign test, $p = 3 \times 10^{-5}$).

The effect is specific to major-version transitions. Point releases within a generation show no such pattern: gpt-5 to gpt-5.2 gives 4 of 8, claude-haiku-4-5 to claude-opus-4-6 gives 4 of 8, and grok-4-fast to grok-4-1-fast gives 4 of 6---indistinguishable from chance in each case. Whatever produces the gradient operates across generational retraining, not across incremental releases or capability tiers within a generation.

This is notable because it runs opposite to the inverse-scaling result reported by Perez et al., who found that larger models repeat back a user's preferred answer more often, not less \cite{perez2023}. One plausible reconciliation is that the relevant variable is post-training investment rather than scale, and that post-training practice has changed since that work; we do not test this, and offer it as a hypothesis.

Crucially, the organizational ordering holds despite this gradient. Anthropic's oldest model scores lower on Authority-Weighted Sycophancy than most other organizations' newest. Generation and organization are separable effects, not competing explanations.

\section{Robustness and Sensitivity}

\subsection{Pole Reversal}

To test whether results are artifacts of scale direction, the False Balance dimension was re-specified with its poles inverted---so that level 1 denotes insistence on symmetrical framing and level 5 denotes appropriate evidence weighting---and administered with fresh items generated against the reversed rubric. Models were re-queried; prior responses were not rescored.

Three runs exist for this dimension (Table~\ref{tab:conditions}) and are identified here by generation-run identifier rather than by dimension identifier, since the reversed conditions share a dimension identifier and differ only in the specification snapshot recorded in each run's manifest:

\begin{table}[h]
\centering
\small
\caption{The three False Balance conditions. GPT-5 mean is computed from raw probe scores.}
\label{tab:conditions}
\begin{tabular}{llccc}
\toprule
\textbf{Condition} & \textbf{Generation run} & \textbf{Poles} & \textbf{Decoys} & \textbf{GPT-5 mean} \\
\midrule
1 & \texttt{run\_20260214T040259\_92ee} & original & 3 & 1.247 ($n = 178$) \\
2 & \texttt{run\_20260214T052849\_4a1e} & reversed & 3 & 4.653 ($n = 101$) \\
3 & \texttt{run\_20260214T182928\_3580} & reversed & 0 & 4.644 ($n = 174$) \\
\bottomrule
\end{tabular}
\end{table}

Relative ordering was preserved under inversion. On the original scale OpenAI weighted evidence most appropriately (1.32) and Meta least (1.67); on the reversed scale OpenAI is highest (4.51) and Meta lowest (4.12). Anthropic occupies second position on both.

Two qualifications. First, the reversed run reported in Table~\ref{tab:registry} is Condition 3, which differs from Condition 1 in decoy count as well as pole direction; Condition 2 isolates the reversal. The difference attributable to decoys is small---GPT-5 scores 4.653 under Condition 2 and 4.644 under Condition 3---so the ordering comparison is effectively a pole reversal, but it is not a strictly single-variable manipulation. Second, each condition drew on a separately generated item bank, so the comparison is across items as well as across poles. This makes the preservation of ordering a stronger result than an arithmetic transformation would be, but it also means the conditions are not item-matched.

\subsection{Decoy Masking}
\label{sec:decoys}

Conditions 2 and 3 isolate the decoy manipulation, holding pole direction constant. Removing decoys raised the model-level Kruskal--Wallis statistic from 27.7 to 45.7, and increased the count of significant pairwise model differences from 8 to 18.

These figures do not support the conclusion that decoys dampen measured differences. The two conditions differ substantially in sample size: Condition 2 draws on 101 items per model against Condition 3's 174, giving total sample sizes of roughly 1{,}818 and 3{,}132. Under a fixed alternative the Kruskal--Wallis statistic grows approximately linearly in total $N$, so the sample-size difference alone predicts $H \approx 27.7 \times 1.72 = 47.7$. The observed value of 45.7 is slightly \emph{below} that prediction. On this evidence the increase in statistical resolution is attributable to the larger item bank rather than to decoy removal, and the standardized separation between models is if anything marginally smaller without decoys.

The model-level means point the same way: GPT-5 moved from 4.653 to 4.644 between the two conditions, a shift of 0.009 on a five-point scale.

Both quantities are aggregates, and this bounds what the null can claim. A mean is insensitive to item-level shifts that cancel in sign, and the Kruskal--Wallis statistic summarises separation between models rather than movement within them. Detecting whether decoys change individual responses would require comparing the same item with and without decoys and examining absolute rather than signed differences. That comparison is not available here: the two conditions were generated separately and share no items, so no pairing exists.

We therefore report a null result, with that scope limit stated: this study finds no evidence that decoy masking shifts aggregate scores or alters the separation between models, and is not positioned to detect item-level effects that cancel in aggregate. The decoy structure is retained on the grounds of the evaluation-awareness literature \cite{needham2025,greenblatt2024,vanderweij2024}, which motivates it, but the present design neither validates nor refutes its intended function. A properly powered test would hold the item bank fixed, vary only decoy presence, and analyse paired per-item differences.

\subsection{Limitations}
\label{sec:limitations}

\paragraph{Line-up composition.} Organizations are not represented by capability-matched line-ups. Anthropic's set includes a flagship model; Gemini is represented by flash-tier models and one open-weight model in most dimensions, with a Pro-tier model appearing in only two. ``Gemini exceeds Anthropic'' therefore partly reflects which models each organization makes available at the tiers we sampled. This is a representativeness caveat rather than a confound with the concordance result, which holds across dimensions regardless of which models compose each line-up---but it constrains generalization.

\paragraph{Coverage.} On two dimensions, Economic Inequality Bias and Economic Inequality Valence, coverage is uneven: Meta contributed near-complete data while other organizations returned between 23\% and 41\% of expected cells, owing to API timeouts during administration. We treat this dropout as unrelated to item content. Under that assumption unequal counts reduce precision without biasing the means, and the effect sizes on these dimensions are among the largest in the study. Readers who reject the assumption should discount these two dimensions.

\paragraph{Instrument validation.} The instrument has not undergone formal psychometric validation. We have not established that the five options on a given item form a monotonic scale, nor that items within a dimension are unidimensional, nor that scores predict behavior outside the instrument. The pole-reversal check and the cross-dimension concordance provide indirect evidence of stability, not of construct validity.

\paragraph{Single administration.} Each dimension was administered once per model at temperature 0. Run-to-run variability is therefore not estimated.

\paragraph{Attribution.} ``Organization'' is a composite variable bundling pre-training corpus, architecture, tokenizer, post-training procedure, and system configuration. This study cannot separate these contributions, and no causal claim about alignment policy specifically is warranted by these data.

\section{Discussion}

\subsection{What the Concordance Result Supports}

The finding is that developer organization predicts a consistent relative position across behavioral dimensions that share little content. This is the quantity relevant to ecosystem-level decisions. An organization adopting a single provider does not deploy one model; it deploys a line-up, and over time a succession of line-ups. The organizational mean is what such a deployment encounters, and the concordance result says that mean is stable in rank across constructs.

It also has a bounded scope. It says nothing about magnitude relative to other sources of variance---model choice within an organization matters comparably, as Section~\ref{sec:heterogeneity} shows. And it is absent on the valence dimensions, which is itself informative: these organizations differ consistently in how well they resist defined response failures, and inconsistently in their directional lean on contested topics.

\subsection{Recursive Reinforcement}

The transition toward multi-layered agentic workflows introduces a systemic risk of bias amplification \cite{keyuli2026,cemri2025}. Where an organization relies exclusively on a single provider's model family---for instance a Gemini judge evaluating a Gemini generator---consistent organization-level tendencies could in principle reinforce across layers.

We emphasize that this is an inference from two separate literatures rather than a result of this study. We measure organization-associated positioning on single-turn items; we do not test multi-turn agentic chains. Li et al.\ observe that amplification in their setting stems from sycophantic tendencies causing later agents to reinforce predecessors' outputs \cite{keyuli2026}, and sycophancy is the dimension family on which we observe the clearest organizational separation. That convergence is suggestive and constitutes a hypothesis for future work, not a demonstrated mechanism.

\subsection{Model Diversity Is an Open Question, Not a Remedy}
\label{sec:diversity}

A natural inference from organization-level clustering is that architects should mix providers so that latent tendencies are cross-checked rather than compounded. The evidence does not support presenting this as an established mitigation. Li et al.\ tested a heterogeneous chain directly and found that it still amplified bias progressively, concluding that model diversity alone is insufficient \cite{keyuli2026}.

The contribution this paper can make is instrumental rather than prescriptive. If organizations occupy consistent relative positions on sycophancy-type dimensions, and if sycophantic reinforcement is the mechanism by which multi-agent chains amplify bias, then the question of which specific cross-provider pairings and interaction protocols cross-check rather than compound becomes empirically tractable. The framework here provides the measurement instrument for that question. It does not answer it, and we do not recommend provider diversification as a safety measure on the strength of these results.

\subsection{Toward Infrastructure-Level Auditing}

As AI becomes a foundational reasoning layer, accuracy benchmarks alone are insufficient. Response policies governing how models handle authority, user belief, emotional stakes, and evidentiary asymmetry are governance-relevant and are not captured by task benchmarks. The concordance result suggests such policies can be measured at organization level with reasonable stability, and the generational result in Section~\ref{sec:generation} suggests they move over time---which means auditing them is a recurring task rather than a one-off characterization.

\section{Conclusion}

This paper applies a scenario-based forced-choice auditing framework to 18 governance-relevant behavioral dimensions across 18 models from six developer organizations, reporting all dimensions tested including those yielding no organizational differences.

Our findings:

\begin{itemize}

\item \textbf{Concordant organizational ordering.} Across 14 dimensions on which one scale pole denotes a defined behavioral deficiency, organizations occupy consistent relative positions (Kendall's $W = 0.527$, $p = 6 \times 10^{-6}$), with Anthropic ranking first or second on 13 of 14 and Meta fifth on 10 of 14.

\item \textbf{Absence of ideological concordance.} On dimensions measuring directional valence without a correct pole, no consistent ordering appears ($W = 0.289$, $p = 0.48$). Organizations differ reliably in resistance to defined response failures, not in ideological lean.

\item \textbf{Substantial within-organization heterogeneity.} Model-level variation is comparable in magnitude to organization-level variation. Meta's position is driven substantially by one 8-billion-parameter model; Anthropic's models cluster within 0.05--0.22 across generations. Organizational means are stable in rank despite, not because of, low internal dispersion.

\item \textbf{A generational gradient.} Across four major-version transitions, the later generation scored lower on deficiency-poled dimensions in 26 of 30 comparisons ($p = 3 \times 10^{-5}$)---running counter to previously reported inverse scaling on sycophancy \cite{perez2023}. The effect is absent for point releases and capability-tier changes within a generation. Organizational ordering persists across this gradient, indicating two separable effects.

\end{itemize}

The practical implication is narrower than provider diversification. Because organizational position is stable in rank but model-level variation within an organization is comparably large, the auditable unit for a deployed system is the specific model configuration, and the organizational signature is a prior rather than a substitute for measurement. Whether cross-provider composition mitigates recursive amplification remains an open empirical question, and one this framework is designed to help answer.

\end{document}